\documentclass[sigconf]{acmart}

\AtBeginDocument{%
  }

\setcopyright{rightsretained}
\copyrightyear{2025}
\acmYear{2025}
\acmDOI{10.1145/3712256.3726329}
\acmConference[GECCO '25]{Genetic and Evolutionary Computation Conference 2025}{July 14--18, 2025}{Málaga, Spain}
\acmISBN{979-8-4007-1465-8/2025/07}
\usepackage{xcolor}
\usepackage{xspace}

\newcommand{\implacro}{M2N2\xspace}
\begin{document}

\title{Competition and Attraction Improve Model Fusion}

\author{João P. Abrantes}
\author{Robert Tjarko Lange}
\author{Yujin Tang}
\affiliation{%
  \institution{Sakana AI}
  \city{}
  \country{}
}
\email{{joao,robert,yujintang}@sakana.ai}

\begin{abstract}
Model merging is a powerful technique for integrating the specialized knowledge of multiple machine learning models into a single model.
However, existing methods require manually partitioning model parameters into fixed groups for merging, which restricts the exploration of potential combinations and limits performance.
To overcome these limitations, we propose Model Merging of Natural Niches (\implacro), an evolutionary algorithm with three key features:
(1) dynamic adjustment of merging boundaries to progressively explore a broader range of parameter combinations;
(2) a diversity preservation mechanism inspired by the competition for resources in nature, to maintain a population of diverse, high-performing models that are particularly well-suited for merging;
and (3) a heuristic-based \textit{attraction} metric to identify the most promising pairs of models for fusion.
Our experimental results demonstrate, for the first time, that model merging can be used to evolve models entirely from \textit{scratch}.
Specifically, we apply \implacro to evolve MNIST classifiers from scratch and achieve performance comparable to CMA-ES, while being computationally more efficient.
Furthermore, \implacro scales to merge specialized language and image generation models, achieving state-of-the-art performance.
Notably, it preserves crucial model capabilities beyond those explicitly optimized by the fitness function, highlighting its robustness and versatility.
Our code is available at \url{https://github.com/SakanaAI/natural_niches}.
\end{abstract}

\maketitle

\section{Introduction}

Open-source generative models have enabled the proliferation of thousands of specialized variants, fine-tuned by practitioners to meet their specific needs.
In an environment where diverse models are freely accessible, the ability to merge and consolidate this wealth of knowledge into a single model becomes increasingly valuable.
This process, known as model merging~\cite{Labonne2024}, has gained traction, as evidenced by the widespread presence of merged models on the Open LLM Leaderboard~\cite{HuggingFace2023}.

\begin{figure}[!t]
    \centering
    \vspace{5mm}
    \includegraphics[width=0.48\textwidth]{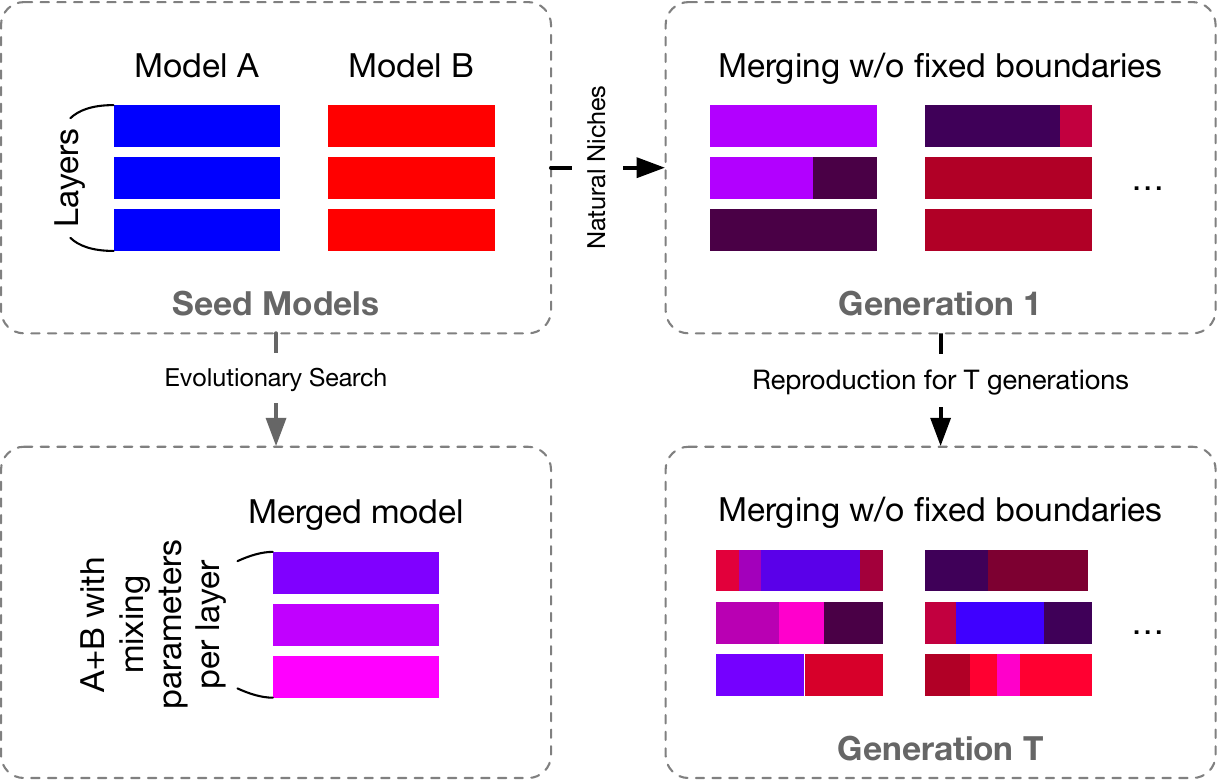}
    \caption{Left, previous methods group the parameters of each seed model according to fixed boundaries (e.g., model layers) and then search for a set of coefficients to mix each group. The shades of purple in the layers of the merged model represent how much the interpolation is close to parent A (blue) or parent B (red). Right, evolution of an archive of models using a random split-point explores a progressively larger number of coefficients and boundaries.}
  \label{fig:splitpoint}
  \vspace{-5mm}
\end{figure}

Model merging initially relied on manually adjusting coefficients to combine seed models, a process guided by intuition and requiring significant trial and error to optimize performance for specific tasks.
Recently, evolutionary algorithms have streamlined this process by automatically searching for optimal coefficients~\cite{akiba2024evolutionary,kuroki2024agent}, significantly improving merging efficiency.
However, one manual step remains: developers must group model parameters into fixed sets before merging, restricting the search space for potential combinations (see Figure~\ref{fig:splitpoint} left).
To address this limitation, we propose Model Merging of Natural Niches (\implacro), an evolutionary algorithm with three key features:

\begin{enumerate}
    \item \textbf{Evolving the Merging Boundaries}. Existing methods partition the parameters of each seed model into fixed groups (e.g., layers) and optimize merging coefficients within these predefined boundaries, limiting the scope of exploration. In contrast, \implacro iteratively merges two models at a time, using flexible split points to divide parameters. Instead of working with static models, we maintain an evolving archive of models. As the number of generations increases, \implacro progressively explores a broader set of boundaries and coefficients (see Figure~\ref{fig:splitpoint} right), enabling increasingly complex combinations when beneficial. This gradual, optional increase in complexity ensures a more extensive search while maintaining computational efficiency.
    \item \textbf{Managing Diversity}. Model merging is most effective when combining diverse models, making diversity preservation essential. However, the challenge lies in defining which characteristics should remain diverse. Many approaches require manually specified diversity metrics, however, we believe it is becoming increasingly hard to come up with efficient diversity metrics as models and tasks grow in complexity. Instead, we incentivize diversity of high-performing models as Nature does --- through competition for limited resources. 
    
    \item \textbf{Attraction}. We introduce a heuristic for pairing models based on their complementary strengths, which improves both efficiency and the final model performance. \textit{Mate selection} remains an underexplored aspect of genetic algorithms, yet it becomes increasingly crucial as the computational costs of crossovers (merging) grow. \implacro highlights the importance of this factor and encourages further research in this area.
\end{enumerate}

Our achievements and key contributions are summarized below.
\begin{itemize}
    \item We introduce \implacro, a novel evolutionary approach with three key components -- competition, attraction, and model fusion with split points. Through comprehensive ablation studies, we demonstrate that these components significantly improve model merging and could enhance other evolutionary algorithms using crossover operations.
    \item We present the first application of model merging for training models from scratch, showing that our method surpasses existing evolutionary algorithms in both performance and computational efficiency.
    \item We scale \implacro to Large Language Models (LLMs) and diffusion-based image generation models, highlighting key benefits of gradient-free optimization, including stable model fusion without catastrophic forgetting, compatibility across models trained on different objectives, reduced memory footprint by avoiding gradient computations, and preservation of model capabilities without requiring access to the original training data.
\end{itemize}

\section{Related Work}

In this section, we review two relevant areas of research: the emerging field of model merging, and the well-established diversity preservation mechanisms in genetic algorithms. While model merging represents a novel approach that has evolved rapidly from manual to automated techniques, diversity preservation in evolutionary algorithms has been extensively studied over several decades.

\subsection{Model Merging}
Model merging introduces an innovative approach for integrating the strengths of multiple pre-trained models. In contrast to fine-tuning, which focuses on refining a single pre-trained model, model merging can leverage several models concurrently without requiring back-propagation. This has allowed the method to combine extremely large models for tasks involving subjective goals, like customizing an image generation model to reflect personal tastes. Notably, the release of Stable Diffusion (SD) \cite{rombach2022high} and open-source interfaces \cite{automatic1111_2022} enabled practitioners to merge different SD fine-tunes manually, using techniques like linear and spherical linear interpolation (SLERP) \cite{white2016sampling}. These early efforts demonstrated the potential of model merging in combining specialized capabilities into a single unified model. Curiously, this modern trial-and-error search for personally appealing images seems extremely inline with earlier evolutionary art systems like Picbreeder \cite{secretan2008picbreeder}, where users could combine and evolve neural networks to create images that matched their subjective preferences through an exploratory interface.

Subsequent research has approached the model merging problem from two complementary directions: minimizing interference between models and automating the merging process. Methods such as TIES \cite{yadav2023tiesmergingresolvinginterferencemerging} and DARE \cite{yu2024language} introduced strategies to balance the contributions of individual models while minimizing interference, ensuring that the strengths of each model are preserved without mutual disruption. They also expanded the applications of model merging to the natural language domain.

Evolutionary algorithms like CMA-ES \cite{hansen2001completely} were later applied to automate the search for optimal merging coefficients. As explored in \cite{akiba2024evolutionary}, these methods not only automate what was previously a manual, iterative process but also significantly improve efficiency.

While previous research centered on merging pre-trained models, we show that merging can efficiently be used to evolve models from scratch. Additionally, unlike earlier methods that required manual partitioning of model parameters, we automate and optimize this process during the evolutionary process.

\subsection{Overview of Diversity Preservation in Genetic Algorithms}
Diversity preservation in Genetic Algorithms (GA) is crucial for finding multiple solutions to multimodal problems \cite{wong2015evolutionary} and to prevent premature convergence. We believe this is particularly important when using cross\-over operations (such as model merging). These operations benefit from diversity while at the same time reducing it, which may lead to premature convergence if not counter-acted by a diversity increasing mechanism. In this section, we provide a quick overview of the two main methods for diversity preservation in GA: 1) crowding \cite{de1975analysis, wong2012evolutionary} and 2) fitness sharing \cite{goldberg1987genetic, deb1989investigation, goldberg1992massive, petrowski1996clearing}.

\textbf{Crowding methods} involve first applying mutation and crossover to produce new candidate solutions. These candidates then compete for inclusion in the population, but only against other candidates that are similar, based on a predefined criteria such as genetic or phenotypic distances. This selective competition helps maintain diversity within the population by preventing any single solution type from becoming overly dominant. A similar mechanism for selective competition is used in the popular algorithm of MAP-Elites \cite{mouret2015illuminating}. In MAP-Elites, the solution space is divided into a multidimensional grid, with each cell representing a species defined by one or more predefined behavior descriptors. New candidates are placed into cells based on their descriptors and replace existing solutions only if they perform better. The real challenge of this method lies in defining descriptors that promote the desired type of diversity.

\textbf{Fitness sharing} requires each individual to share its rewards with others. In \emph{explicit} fitness sharing, the researcher defines a distance function that is used to cluster similar individuals into a species, each individual then shares its fitness with other members of its species, making it more difficult for any single species to grow excessively large. A notable example is the NEAT \cite{stanley2002evolving} algorithm, known for evolving neural networks topologies, which clusters solutions into species by measuring genotypic differences (distances in network topologies). \textit{Implicit} fitness sharing \cite{smith1993searching, darwen1996every}, is seen as the more natural method because, as in Nature, it protects niches rather than species. A niche is a group of individuals that compete for the same resources, while a species is defined as group that can interbreed and typically have small genetic and phenotypic differences. Usually, members of the same species compete for the same resources (e.g., food, partners, shelter), but vastly different species can also compete for the same vital resources like nesting sites or food sources (e.g., birds and bats, lions and hyenas). \textit{Implicit} sharing does not rely on custom distance metrics. Instead, it simulates natural competition for limited resources, promoting diversity as individuals who can derive their fitness from less contested resources are favored. While this approach has received less attention recently, we argue it deserves renewed consideration. The high-dimensional nature of modern AI tasks and models makes defining meaningful diversity metrics particularly challenging. By leveraging natural resource competition rather than explicit distance metrics, implicit fitness sharing offers an elegant and scalable solution for contemporary applications. We hope this work helps revitalize interest in this underappreciated approach to diversity maintenance. We provide more details in Section \ref{sec:method}.

\section{Natural Niches}
\label{sec:method}

In model merging, the goal is to find the optimal parameters $\theta^*$ for a merged model from a set of $K$ seed models, each of which is characterized by its model parameters $\theta_i$ ($i=1 \cdots K$), so that the optimization goal, normally in the form of summation or average of task scores, is maximized. The following equation expresses this description mathematically:
\begin{equation}
    \theta^* = \arg\max_{\theta} \sum_{j=1}^N{s(x_j \mid \theta}), \text{where}, \theta = h_w(\theta_1, \cdots, \theta_K)
\label{eq:goal}
\end{equation}
Here, $h_w$ is the model merging function parameterized by $w$'s that correspond to fixed model merging boundaries (e.g., one scalar $w_{k,l}$ for the $l$-th layer in the $k$-th seed model), $s$ is the score function for a certain task, $x_j$ is a task example, and $N$ is the number of examples to be evaluated.
In Natural Niches (\implacro), we propose modifications to the merging function $h$ to allow the evolution of the merging boundaries, and adjustments to the optimization goal to promote diverse solutions.

\subsection{Eliminating Fixed Model Merging Boundaries}
\label{sec:method_merging}

In the formulation above, finding $\theta^*$ boils down to searching for the optimal model merging parameters $w$ in $h_w$.
To get rid of the constraints of fixed model merging boundaries and thus allow more flexibility, we propose to include these boundaries together with the mixing parameters into the evolutionary process. 
Concretely, \implacro maintains an archive of models, which is initialized with the $K$ seed models. At each training step, \implacro randomly picks two models $A$ and $B$ from the archive, and samples two parameters $(w_m, w_s)$ that determines the mixing ratio and the split-point in the models' parameters space.
It then merges models $A$ and $B$ with the following formula, and inserts the new model into the archive if it outperforms the worst individual.
\begin{equation}
\begin{aligned}
    h_\mathrm{\implacro}(\theta_A, \theta_B, w_m, w_s) = \text{concat}\big( & f_{w_m}(\theta_A^{<w_s}, \theta_B^{<w_s}), \\
    & f_{1-w_m}(\theta_A^{\ge w_s}, \theta_B^{\ge w_s}) \big)
\end{aligned}
\label{eq:merge_with_splitpoint}
\end{equation}
Here, $\theta^{<w_s}$ and $\theta^{\ge w_s}$ indicate the sub-arrays of model parameters before and after the split-point indexed by $w_s$. $f_t(\theta_A, \theta_B)$ is a spherical linear interpolation of rotations (SLERP) function that interpolates $(\theta_A, \theta_B)$ with $t$.
As shown in the right part of Figure~\ref{fig:splitpoint}, our method incrementally expands the search space by exploring a broader set of boundaries and coefficients. This gradual introduction of complexity ensures a wider range of possibilities while maintaining computational tractability.

\subsection{Encouraging Diversity via a Modified Optimization Goal}

Competing for limited resources \textbf{naturally} promotes diversity, favoring individuals who can tap into less contested resources. In the context of the optimization goal in Equation~\ref{eq:goal}, where a sum of scores from all the examples is being maximised, each score is a ``resource'' that contributes to the fitness of a solution. By limiting the resource supply, \implacro sparks competition which naturally favors individuals that take over new niches. Concretely, we limit the total fitness a population can extract from a data point $x_j$ by a capacity $c_j$. The amount of fitness a candidate solution derives from a data point is proportional to its score relative to the aggregate score of the population. Our modified goal becomes:

\begin{equation}
        \theta^* = \arg\max_{\theta} \sum_{j=1}^N{\frac{s(x_j \mid \theta)}{z_j+\epsilon}c_j}, \text{where}, z_j = \sum_{k=1}^{P}s(x_j \mid \theta_k)
\label{eq:nn_fitness}
\end{equation}

where $\epsilon$ in the denominator is a small number to prevent the zero-division error. In the term that defines $z_j$, $P$ is the archive size. The capacity $c_j$, is task dependent and can be defined in multiple ways.
For example, in binary scoring tasks (i.e. $s(\cdot) \in \{0, 1\}$) we simply set $c_j=1$. In some experiments, we have a continuous reward from 0 to 1. Here, we define $c_j=\max_i s(x_j|\theta_i)$ to ensure that partially solved data points (where $\max_i s(x_j|\theta_i) < 1$) do not distribute the same amount of fitness points as fully solved data points (where $\max_i s(x_j|\theta_i) = 1$).

\subsection{Sampling Parents via Attraction}
\label{sec:matchmaker}

Many evolutionary algorithms use the crossover operation to combine the strengths of both parents. In biology, this combination (i.e., reproduction) is very expensive, and therefore, animals invest many resources in the process of mate selection. We believe that as we make use of more expensive crossover operations, like model merging, algorithms for mate selection become increasingly important.

In contrast to conventional methods that put more sampling probability mass on top performing models in the archive, \implacro adds an extra layer of consideration that takes into account the complementarity of the parent models.
Specifically, we sample the first parent based on their weighted sum of scores defined in Equation~\ref{eq:nn_fitness}, and then sample the second parent based on a ``attraction score'' generated by function $g$ that is specifically tailored for the first parent.
The equation below gives the definition of this attraction score, it straightforwardly expresses a desire to choose a model B that performs well in the data points where model A performs less well, while giving an extra preference to resources with high capacity $c_j$ and low competition $z_j$.
\begin{equation}
    g(\theta_A, \theta_B)=\sum_{j=1}^N{\frac{c_j}{z_j+\epsilon} \max \big( s(x_j \mid \theta_B)-s(x_j \mid \theta_A), 0 \big)}
\label{eq:matchmaker}
\end{equation}

\section{Experiments}

We verify the effectiveness of our proposed method on three challenging tasks: First, we evolve image classifiers from scratch and from pre-trained models on the MNIST dataset, then scale up the experiment to merging LLMs and diffusion-based image generation models to demonstrate its general applicability.

\subsection{Experiment 1: Evolving MNIST classifiers}
\label{exp:mnist}


\vspace{1mm}
\noindent\textbf{Setup}
\vspace{1mm}

\underline{Model}: The model being optimized is a two-layer feedforward neural network with 19,210 parameters in total. When starting from scratch, we randomly initialize the models. For pre-trained models, we develop two specialized models: one is trained on digits 0 through 4, and the other is trained on digits 5 through 9.

\underline{Baselines}: For the MAP-Elites algorithm, we use two diversity dimensions to create a 10 by 10 grid: the accuracy of the model in odd and even numbers. When starting from scratch, we use CMA-ES \cite{hansen2001completely} as a baseline, even though it does not perform model merging here. Since the models are randomly initialized, optimizing mixing coefficients alone would be insufficient. Instead, CMA-ES directly optimizes model weights, which incurs a cubic computational cost $O(n^3)$ with respect to the number of parameters. While this method doesn't scale to larger models, it serves as a benchmark for how a popular evolutionary algorithm performs in this experiment. When working with pre-trained models, we use a brute-force search baseline that merges the two seed models by adjusting a mixing coefficient that ranges from 0 to 1 in increments of $10^{-5}$. This baseline is first evaluated on the training data; the best coefficient is subsequently evaluated on the test data.

\underline{Evolutionary Operators and Variables}: All model merging methods (which excludes CMA-ES) sample a new candidate at a time and decide sequentially whether to insert the candidate into the archive. \implacro and GA use an archive of 20, sampling each candidate sequentially and deciding whether to insert it, similar to MAP-Elites. MAP-Elites, uses a 10x10 grid, resulting in an archive size of 100. CMA-ES uses a population of 20, sampling and updating its parameters in batches. When starting from scratch all model merging methods use the same mutation operation (Gaussian noise) and the same crossover operation (SLERP with split-point, as described in section \ref{sec:method_merging}). However, when dealing with pre-trained models, mutation is omitted because we want to assess how our method would scale to larger models where random mutations are not effective.

\underline{Compute Resources}: The 10 independent runs took about 15 hours for CMA-ES, and about 1h for each of the other methods. We ran this experiment using only CPUs.

\begin{figure}[!t]
    \centering
    \includegraphics[width=0.48\textwidth]{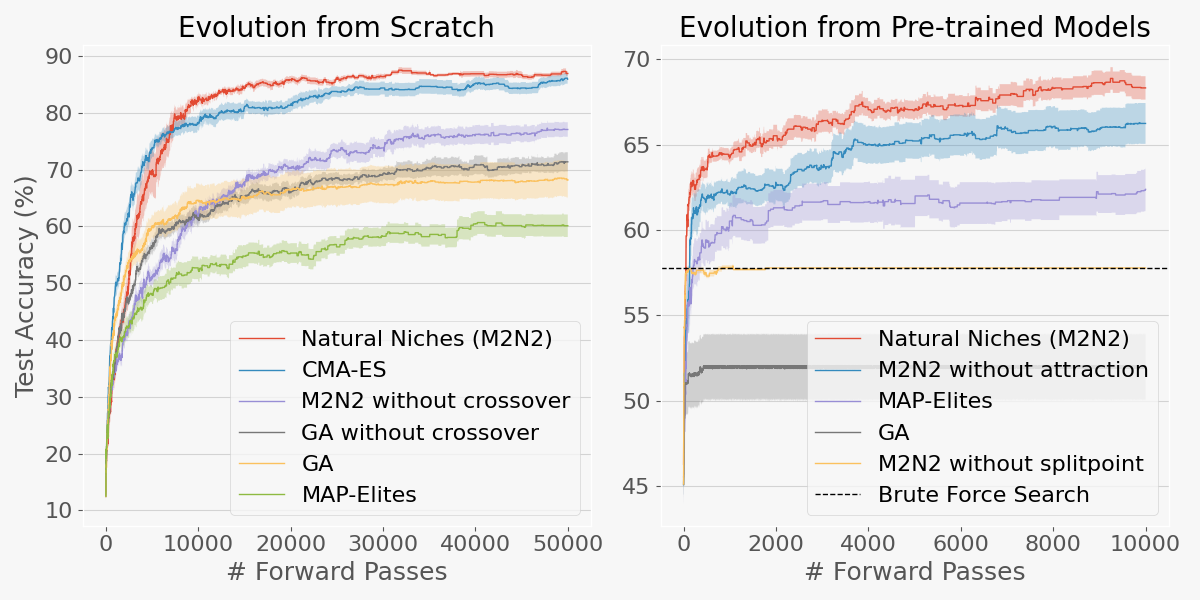}
    \caption{The plots show the accuracy on the test split vs the number of forward passes when starting from randomly initialised models (left) and when starting from the pre-trained models (right). The solid-lines represent the mean of ten independent runs and the shaded area around represents one standard error deviation.}
    \label{fig:mnist_results}
    \vspace{-5mm}
\end{figure}

\vspace{1mm}
\noindent\textbf{Results}
\vspace{1mm}

When starting from scratch, \implacro achieves the highest test accuracy by a substantial margin when compared to the other model merging methods, as shown in Figure \ref{fig:mnist_results} (left). Interestingly, GA with crossover performs better early on (before step 12,000) than GA without crossover, however, it converges faster to an inferior solution. The early convergence happens because GA can't maintain a diverse population which is crucial for effective crossover operations. Crossover reduces population diversity, and without a strong counteracting force, it diminishes exploration. In contrast, \implacro leverages the crossover operation effectively, benefiting significantly from the diversity it manages to retain. GA is an extreme case where there is no competition, we observed that by progressively decreasing competition in \implacro, we progressively converge earlier to worse solutions (analyzed on the next section). MAP-Elites clusters individuals by their accuracy on odd and even numbers. This means it will always keep individuals who perform poorly on those tasks because there is a slot reserved just for them. Even though those individuals add to the diversity of the population, this is clearly not the type of diversity that leads to strong solutions and it highlights the difficulty of hand-engineering useful diversity metrics. 

For models trained from scratch, the split-point and attraction score have a minimal impact (ablations omitted for clarity). However, as seen in Figure \ref{fig:mnist_results} (right), the split-point becomes crucial when starting from pre-trained models, while attraction significantly improves performance throughout the training process. GA has a low average test accuracy with large error bars as its performances is highly dependent on the quality of the first merges. Note that when starting from pre-trained models the mutation operator was not used (as explained in the Setup section), and therefore, the performance is worse.

\vspace{1mm}
\noindent\textbf{Analysis}
\vspace{1mm}

This section focuses exclusively on the experiment where models were evolved from scratch, as the later sections will provide ample discussion on evolving models starting from pre-trained LLMs and diffusion models.

\begin{figure}[!t]
    \centering
    \includegraphics[width=0.48\textwidth]{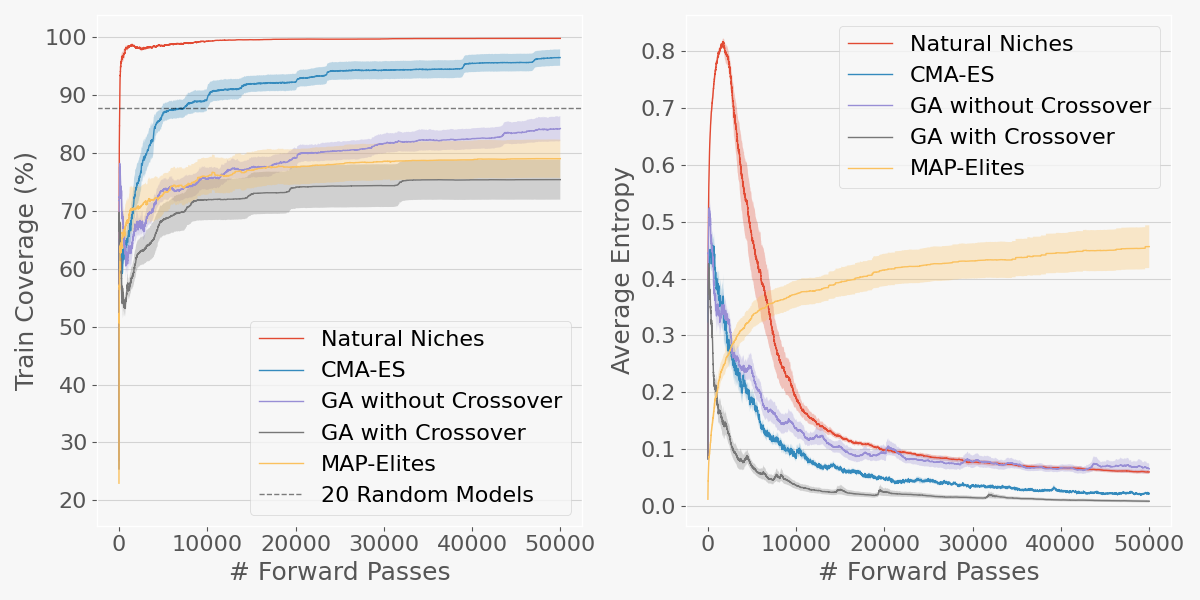}
  \caption{Left: The percentage of training data points that can be correctly labeled by at least one model in the population. Since there are 10 possible labels, 20 random models obtain an average coverage of $1-(\frac{9}{10})^{20}=87.8\%$. Right: The evolution of diversity in the population’s performance, measured by entropy, over the course of training.}
  \vspace{-4mm}
  \label{fig:union}
\end{figure}

\underline{Diversity}: Figure \ref{fig:union} left, shows the percentage of training data points that can be correctly labeled by at least one model in the archive, we call this percentage the training coverage. We observe that the archive in \implacro quickly spreads to cover the majority of the training data points and maintains this high coverage throughout the training process. The right-hand plot shows how the diversity in the performance of the population evolves with training. If either all models correctly or incorrectly classify a data point, the entropy is 0 (no diversity). In contrast, when the models are evenly split on a prediction, entropy reaches its maximum value of 1. The plot displays the average entropy across all data points. For \implacro we see a sharp initial rise in entropy followed by a gradual decline as low-performing models go extinct. In contrast, MAP-Elites continually increases diversity by retaining lower-performing models, but it fails to achieve a high coverage. The Genetic Algorithms, lacking a diversity preservation mechanism, reduce coverage early on and show a sharp drop in entropy as they converge prematurely on the best solutions. This drop in diversity is accentuated with the crossover operation.

Overall, the graphs show that \implacro maintains an archive of models with complementary strengths that facilitate effective merging, while systematically discarding weaker models as training progresses.

\begin{figure}[!t]
    \centering
    \includegraphics[width=0.48\textwidth]{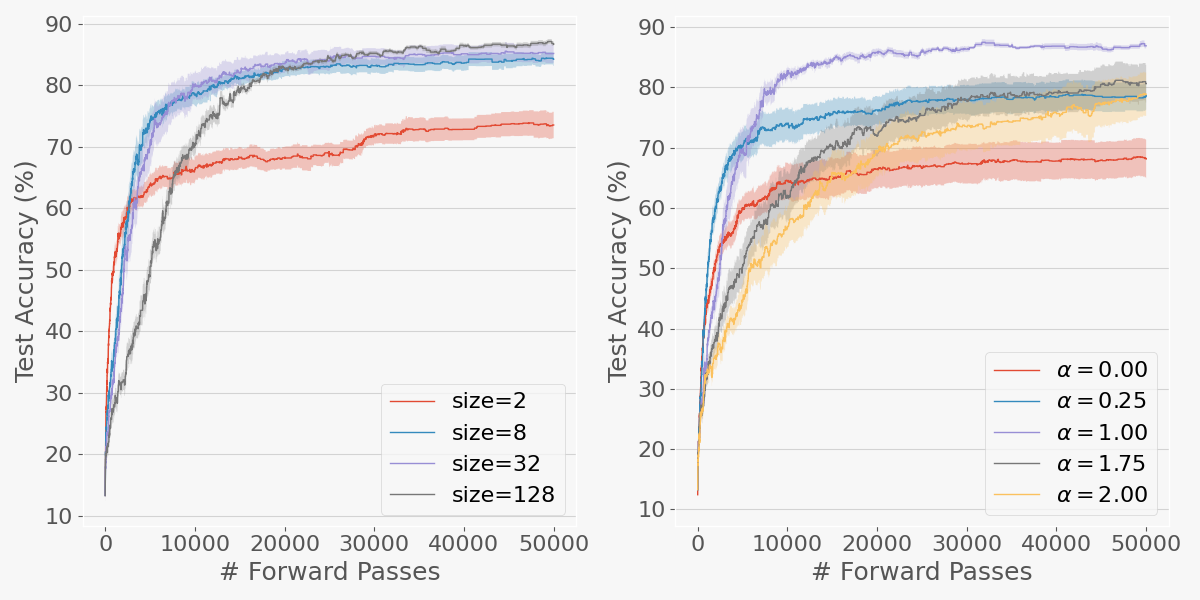}
    \caption{Left, test accuracy of \implacro on the MNIST across different archive sizes. Right, test accuracy for different $\alpha$ values while population size is 20.
}
    \label{fig:population_size}
    \vspace{-5mm}
\end{figure}

\underline{Competition}: Figure \ref{fig:population_size} left, shows that smaller archives perform better in the beginning but converge faster to inferior solutions. This suggests that we should scale the archive size along the number of forward passes we want to make. Note that in our plot the computational cost does not increase with the archive size since the number of forward passes remains the same, however, the memory footprint does increase with larger populations. For very large models we can always store the archive on disk instead of keeping them all in the RAM.

For a fixed population size $P$, we can adjust the intensity of competition by introducing a hyper-parameter $\alpha \geq 0$, as described in the fitness function in eq. \ref{eq:alpha}. 

\begin{equation}    
f(\theta_i) = \sum_{j=1}^N \frac{s(x_j|\theta_i)}{{z_j}^\alpha + \epsilon}c_j
\label{eq:alpha}
\end{equation}

When $\alpha=0$, there is no competition because the total fitness available per data point becomes unlimited. When $\alpha=1$, the total fitness distributed among different individuals is limited to the capacity $c_j$. For $\alpha>1$, the total fitness distributed decreases with increasing competition ($z_j$), this scenario can be thought of as individuals needing to "fight" for resources, spending some fitness points in the process. Figure \ref{fig:population_size} right, shows that smaller values of $\alpha$ (i.e. lower competition) have a similar effect to decreasing the population size: it performs better in the beginning but it converges faster to inferior solutions.

\subsection{Experiment 2: Combining LLMs with Math and Agentic Skills}
\label{exp:llm_merging}

\vspace{1mm}
\noindent\textbf{Setup}
\vspace{1mm}

\underline{Models}: We combine a math specialist, \texttt{WizardMath-7B-V1.0} \cite{luo2023wizardmath}, with a specialist on agentic enviroments, \texttt{AgentEvol-7B} \cite{xi2024agentgym}, to achieve an agent that performs well on the math benchmark GSM8k \cite{cobbe2021training} and on the web shopping benchmark WebShop \cite{yao2022webshop}. These two models share the same architecture of \texttt{Llamma-2-7b} \cite{touvron2023llama}, a decoder-only transformer with 32 layers.

\underline{Datasets}: For the math task, we use the test split of GSM8k as our test split (1319 samples). For the training split, we use the first 1319 samples of GSM8k train dataset. In the web shopping task, we use the WebShop environment implemented in \cite{xi2024agentgym}. The test split consistent of the first 100 tasks, while the training split were the next 100 tasks. We allow the agents to take up to 7 steps.

\underline{Baselines}: The CMA-ES optimizes 32 mixing coefficients (one for each layer) for a SLERP merge between the two seed models. All methods ran for a 1000 evaluations on the training set. For the MAP-Elites we used two dimensions to create a 4 by 4 grid: the accuracy on the math and on the web shopping training splits.

\underline{Evaluation}: In this experiment, all methods used 1000 evaluations on the training set. \implacro and GA used an archive size of 15. CMA-ES used a population size of 25.

\underline{Evolutionary Operators and Variables}: In our LLM Merging experiment we do not use a mutation operator since random mutations don't work well on large models. Moreover, in these experiments we initialize the \implacro and GA archives with seed models, followed by a short warm-up period (50 iterations or less) where the seed models merge randomly amongst themselves and populate the archive.

\underline{Compute Resources}: For these experiments, we used 4 H100 GPUs to run each method for around 24h.

\vspace{1mm}
\noindent\textbf{Results}
\vspace{1mm}

Table \ref{tab:ws_math} shows that \implacro achieves the highest score. Both the attraction and the split-point techniques play a crucial role, however, the split-point seems to be slightly more important. Note that on Table \ref{tab:ws_math} all algorithms run the same amount of evaluations and used the same merging method (SLERP). When combining the Math and Agentic skills, CMA-ES yielded a low score, likely due to suboptimal parameter partitioning, highlighting the need to include the merging boundaries in the optimization process.

\begin{table}[h]
\centering
\caption{Scores of various methods on math (GSM8k) and web shopping (WebShop) benchmarks. The table shows the average and the standard deviation over three independent runs.}
\begin{tabular}{lccc}
\toprule
\textbf{Methods} & \textbf{GSM8k $\uparrow$} & \textbf{WebShop $\uparrow$} & \textbf{Average $\uparrow$} \\
\midrule
WizardMath 7B v1.0             & 74.22 & 0.00  & 37.11  \\
AgentEvol 7B                   & 6.29  & 88.88 & 47.59  \\
\cmidrule(lr){1-4}
\implacro   & 40.16 ± 0.57 & 86.81 ± 1.13 & \textbf{63.49 ± 0.63} \\
\implacro w/o attraction              & 40.34 ± 0.70 & 85.75 ± 1.61 & 63.04 ± 1.14 \\
\implacro w/o splitpoint              & 34.34 ± 2.16 & 87.95 ± 0.53 & 61.15 ± 1.34 \\
GA                             & 32.20 ± 0.96 & 87.91 ± 0.00 & 60.05 ± 0.48 \\
MAP-Elites                     & 36.72 ± 1.63 & 83.57 ± 5.42 & 60.14 ± 1.96 \\
CMA-ES (SLERP)                 & 47.19 ± 0.80 & 50.93 ± 2.94 & 49.06 ± 1.31  \\
\bottomrule
\vspace{-4mm}
\end{tabular}
\label{tab:ws_math}
\end{table}

\vspace{1mm}
\noindent\textbf{Analysis}
\vspace{1mm}

\begin{figure}[!h]
    \centering
    \includegraphics[width=0.48\textwidth]{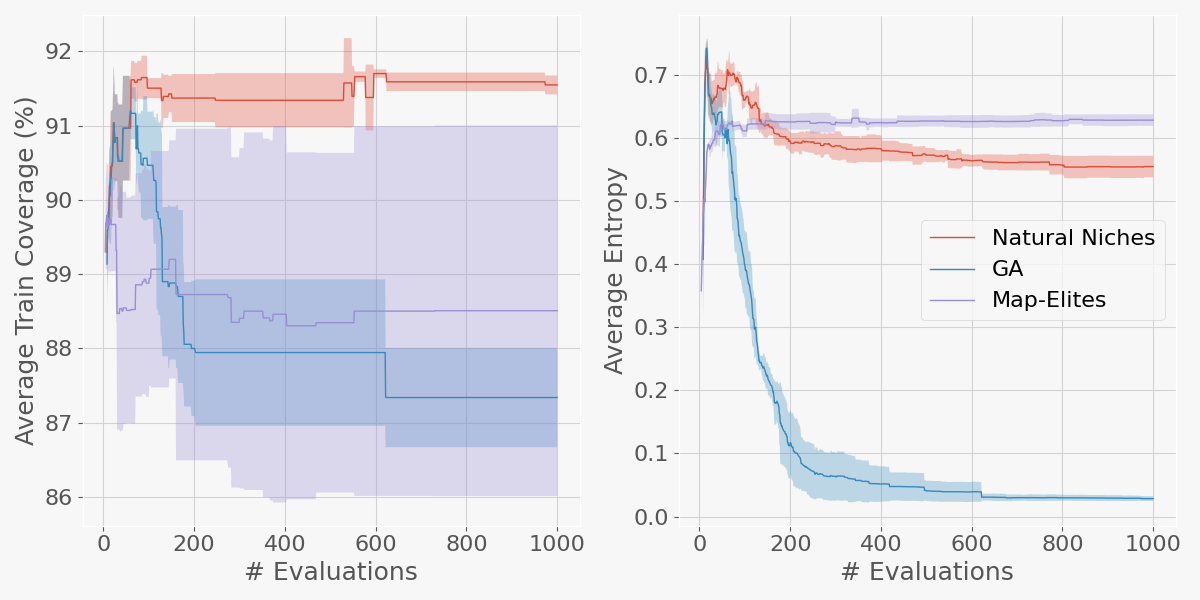}
    \caption{Left, the percentage of training data points that can be correctly labeled by at least one model in the population, averaged on the Math and Web shopping datasets. The right plot shows how the diversity in the performance of the population evolves with training.}
    \label{fig:llm_coverage_entropy}
    \vspace{-5mm}
\end{figure}

As shown in Figure \ref{fig:llm_coverage_entropy}, the findings from the MNIST dataset generalize to LLM merging. The Natural Niches method maintains high training coverage, as seen on the left side of the figure. The entropy rises early on as the models explore diverse niches (right), followed by a gradual decrease as low-performing models are removed, and the strengths of the models are aggregated. In contrast, MAP-Elites focuses on maximizing entropy at the cost of training efficiency and coverage, as it retains low-performing models. GA quickly reduces both coverage and entropy as it greedily converges on its top solution, ultimately collapsing the entire archive onto a single solution, with entropy nearing zero.

\subsection{Experiment 3: Merging Diffusion-Based Image Generation Models}
\label{sec:diffusion}

\vspace{1mm}
\noindent\textbf{Setup}
\vspace{1mm}

\underline{Models}: We evaluate our method in merging diverse text-to-image models. Our seed models include \texttt{JSDXL} \cite{JSDXL}, which was specifically trained on Japanese prompts, and three models primarily trained with English prompts: \texttt{SDXL 1.0} \cite{podell2023sdxl}, \texttt{SDXL-DPO} \cite{wallace2024diffusion}, and \texttt{Juggernaut-XL-v9} \cite{rundiffusion_juggernaut}. All these models share the same architecture of the base model \texttt{SDXL 1.0}. The primary objective is to create a model that combines the best image generation capabilities from each seed model while retaining \texttt{JSDXL's} ability to understand Japanese prompts.

\underline{Evolutionary Operators and Variables}: As in our LLM merging experiment, we omit the mutation operator. For model merging, we retain the VAE from \texttt{SDXL} -- since most seed models left this component unchanged -- and we preserve \texttt{JSDXL}'s tokenizer and text encoder to leverage its superior Japanese language understanding. Therefore, we are combining only the U-Nets from the various models using Equation \ref{eq:merge_with_splitpoint}. We merge the attention layers independently from the other components, as this was found effective on previous work \cite{sakana_evosdxl}. This compartmentalization of parameters is analogous to the grouping of DNA into chromosomes, where each chromosome divides independently and has one or more split points. This design introduces useful inductive biases while maintaining the flexibility to explore more complex merging boundaries.

\underline{Dataset}: We utilize the COCO dataset \cite{lin2014microsoft}, specifically the validation split. We use the last 2,000 images for training and the first 10,000 images for testing, ensuring no overlap between the sets. The Japanese captions are sourced from the STAIR Captions dataset \cite{yoshikawa2017stair}, which provides Japanese descriptions for COCO images.

\underline{Training and Testing metrics}: During training, we give the Japanese captions as prompts to the model and calculate the cosine similarity between the CLIP features of the correspondent images from COCO and the generated ones. This metric is normalized to fall within the range of 0 to 1, which we refer to the Normalized CLIP Similarity (NCS) metric. While models generally achieve high absolute NCS values, small differences can significantly impact performance. To emphasize these relative differences, we adjust the scores before computing the fitness score: for each training sample, we subtract the corresponding worst score achieved by the population. This adjustment emphasizes relative performance, increasing the competition for training samples and favoring more diversity in the population's skills.

\underline{Baseline}: Our work builds on \cite{yoshikawa2017stair} approach using CMA-ES for merging. While we use the same dataset, our method maximizes NCS instead of minimizing the Fréchet Inception Distance (FID) metric \cite{heusel2017gans} as they did. This choice enables per-sample evaluation necessary for resource competition, whereas FID only provides aggregate performance across the training set.

\vspace{1mm}
\noindent\textbf{Qualitative Results}
\vspace{1mm}

\begin{figure}[!h]
    \centering
    \includegraphics[width=0.48\textwidth]{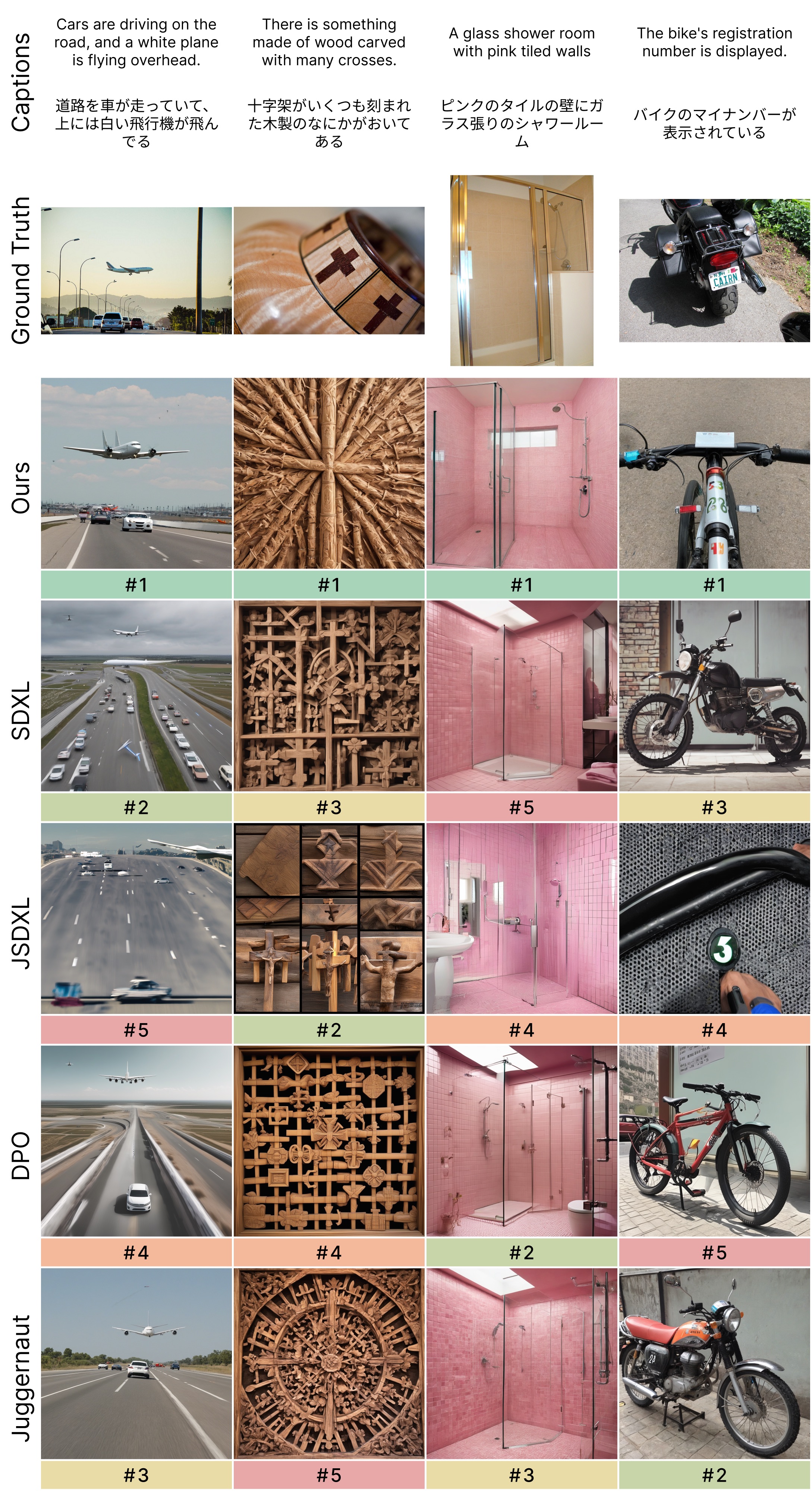}
    \caption{Comparison of generated images across seed models and our merged model, demonstrating performance diversity and successful quality aggregation. Images were prompted with Japanese captions for our merged model and JSDXL, and English translations for other models. Rankings (\#1-\#5) below each image are based on the cosine similarity between CLIP features of generated and ground truth images from the test dataset.}
    \label{fig:aggregate_diverse_skills}
    \vspace{-5mm}
\end{figure}

\underline{Diversity and Visual Capabilities}: Figure \ref{fig:aggregate_diverse_skills} illustrates how our merged model successfully combines the strengths of individual seed models while mitigating their weaknesses. Note that if we exclude our merged model, we observe that each seed model produced both the highest and lowest quality outputs for different test cases. Additionally, it is extremely hard to find a clear pattern that describes the speciality of each model or inform us on how to create an effective and custom diversity metric. Our diversity preservation mechanism addresses this challenge by automatically preserving the models that uniquely excel on training samples where other models underperform.

The merged model demonstrates two key improvements over the seed models. First, it generates more photorealistic images, which aligns with our training set of real photographs. Second, it shows enhanced semantic understanding of the input captions. For instance, in the rightmost column of Figure \ref{fig:aggregate_diverse_skills}, while several seed models generated visually appealing bikes, our merged model not only specifically focused on capturing the bike's registration number display area as specified in the caption, but also produced an image that looks like an actual photograph rather than a synthetic rendering.

\begin{figure}[!h]
    \centering
    \includegraphics[width=0.48\textwidth]{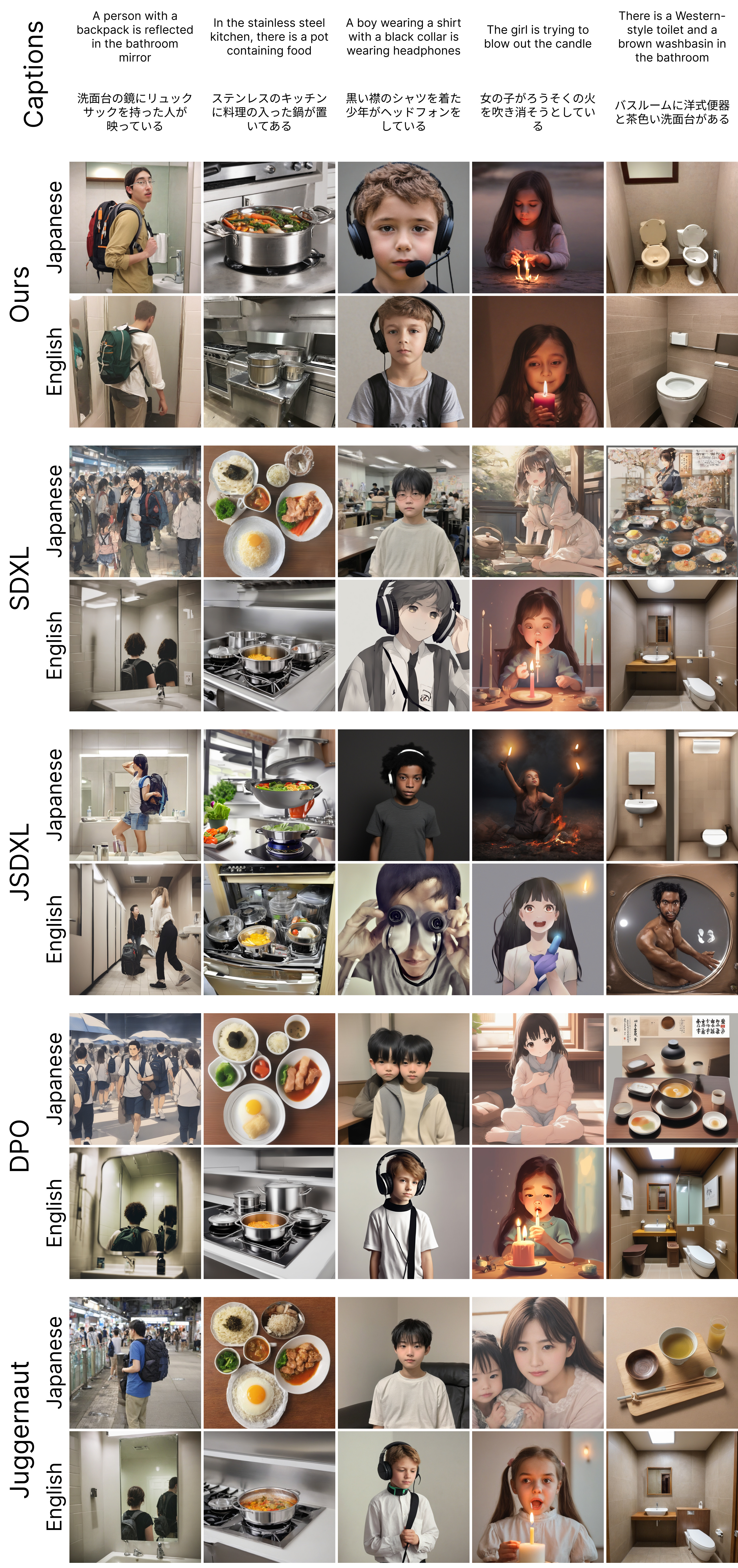}
    \caption{Images generated by each model when receiving Japanese and English prompts.}
    \label{fig:lang_skills}
\end{figure}

\underline{Language Understanding}: Figure \ref{fig:lang_skills} shows that our model has a good understanding of both Japanese and English, despite being evolved exclusively with Japanese captions! This emergent bilingual ability exemplifies a key advantage of model merging: it enables the aggregation of complementary capabilities while avoiding the catastrophic forgetting typically associated with gradient-based training methods. Note that the seed models show very distinct language capabilities: \texttt{JSDXL} performs better with Japanese prompts while the other models perform better with English prompts.

\vspace{1mm}
\noindent\textbf{Quantitative Results}
\vspace{1mm}

\underline{Performance}: Table \ref{table:diffusion} demonstrates that our model achieves a superior NCS score on the test set compared to all other models. Additionally, we surpass the model merging baseline, CMA-ES, in FID performance, even though this baseline was explicitly trained to minimize FID on the training set.

\begin{table}[!t]
    \centering
    \caption{Performance of the various models on the test set.}
    \begin{tabular}{llll}
        \hline
        Model & Type & FID $\downarrow$ & NCS \% $\uparrow$ \\
        \hline
        SDXL 1.0 & Original SDXL & 57.70 & 73.68 \\
        Juggernaut-XL-v9 & Fine-tuned SDXL & 35.56 & 75.80 \\
        SDXL-DPO & RLHF SDXL & 48.54 & 74.13 \\
        JSDXL & JA SDXL & 18.20 & 83.97 \\
        \midrule
        CMA-ES & Model Merging & 13.51 & 84.67 \\
        \implacro & Model Merging & \textbf{13.21} & \textbf{84.85} \\
        \hline
    \end{tabular}
    \vspace{-4mm}
    \label{table:diffusion}
\end{table}

\underline{Language Understanding}: We evaluated the models' ability to maintain semantic consistency across languages by generating pairs of images: one from Japanese captions (sampled from our test set) and another from their \texttt{GPT-4o} translated English versions. For each image pair, we computed the cosine similarity between their CLIP feature representations, which captures how semantically similar the generated images are. Averaging these similarities across 100 diverse caption pairs, Table \ref{table:lang_skills} demonstrates that our model achieves significantly better cross-lingual consistency than the other models. This statistically confirms what we had already observed in our qualitative results.

\begin{table}[h]
\centering
\caption{The average cosine-similarities between the CLIP features of images generated from Japanese prompts and their corresponding English translations, measured across 100 pairs. The reported margins represent standard errors. Higher similarities indicate better cross-lingual consistency.}
\begin{tabular}{lc}
\hline
\textbf{Model} & \textbf{CLIP cosine-similarity $\uparrow$} \\
\hline
\implacro & \textbf{0.787} ± 0.010 \\
JSDXL & 0.701 ± 0.011 \\
Juggernaut-XL-v9 & 0.584 ± 0.014 \\
SDXL-DPO & 0.577 ± 0.013 \\
SDXL 1.0 & 0.567 ± 0.013 \\
\hline
\end{tabular}
\vspace{-2mm}
\label{table:lang_skills}
\end{table}

\section{Limitations \& Future Work}
The feasibility of model merging strongly depends on the degree of similarity between models. As demonstrated in \cite{yu2024language}, when fine-tuned models deviate significantly from their base models—often due to extensive, divergent training—merging becomes impractical. We hypothesize that models with divergent \emph{state representations} are incompatible for merging. However, a standardized metric for model \emph{compatibility} has yet to be established. Defining such a metric could allow it to be used as a form of regularization during preprocessing (e.g., fine-tuning), enabling better control over model compatibility and ensuring the success of merging.

We believe there is a strong evolutionary pressure for models that are co-evolving together to remain compatible for merging. Should one model, diverge and become incompatible with others, it would no longer produce viable offspring, halting its improvement and leading to its eventual extinction. Testing this hypothesis through further research would provide valuable insights into the dynamics of model co-evolution. Moreover, incorporating a \emph{compatibility} metric into the attraction heuristic could facilitate the co-evolution of distinct \emph{species} of models, defined as groups that merge with one another but not with others.

\section{Conclusion}
In this paper, we present the first application of model merging for training models from scratch and demonstrate that it achieves top-tier performance and efficiency when combined with a diversity-preservation technique.
Furthermore, this approach scales effectively to LLMs and diffusion-based image generation models.
Our ablation studies reveal that the proposed mechanisms of competition, attraction and the use of split-points significantly enhance the performance of model merging and have the potential to benefit other evolutionary algorithms utilizing crossover operations.

Our LLM merging experiments highlight the ability to combine vastly different skills, enabling multi-task capabilities without requiring access to the original training data.
Remarkably, in diffusion model experiments, we observe that the merged models retain English language capabilities despite being optimized exclusively for Japanese tasks.
This finding underscores the promise of model merging as a robust transfer learning mechanism that resists catastrophic forgetting, which is typically encountered in fine-tuning.

Finally, we hope this work revitalizes interest in two underexplored areas: mate selection algorithms and implicit fitness sharing as a mechanism for diversity preservation.
These concepts are increasingly critical as crossover operations, such as model merging, become computationally more expensive, and as models and tasks grow in complexity.

\bibliographystyle{ACM-Reference-Format}
\bibliography{library.bib}

\end{document}